\documentclass[a4paper,twoside]{article}

\usepackage{url} 
\usepackage{epsfig}
\usepackage{subcaption}
\usepackage{calc}
\usepackage{amssymb}
\usepackage{amstext}
\usepackage{amsmath}
\usepackage{amsthm}
\usepackage{multicol}
\usepackage{pslatex}
\usepackage{apalike}
\usepackage{algorithm2e}
\usepackage[bottom]{footmisc}
\usepackage{adjustbox} %% PJB
\usepackage{flushend} %% even layout on the last page, PJB
\usepackage{ctable} %% AM for thicker table line
\usepackage{SCITEPRESS} 
\begin{document}

\title{Using Machine Learning to Distinguish 
Human-written from Machine-generated Creative Fiction}

\author{\authorname{Andrea Cristina McGlinchey\sup{1} and 
Peter J Barclay\sup{2}\orcidAuthor{0009-0002-7369-232X}}
\affiliation{\sup{1}Lumerate (https://www.lumerate.com)}
\affiliation{\sup{2}School of Computing, Engineering \& the Built Environment, Edinburgh Napier University, Scotland}
\email{andrea.mcglinchey@lumerate.com, p.barclay@napier.ac.uk}
}


\abstract{
Following the universal availability of generative AI systems with the release of ChatGPT, automatic detection of deceptive text created by Large Language Models has focused on domains such as academic plagiarism and ``fake news''. However, generative AI also poses a threat to the livelihood of creative writers, and perhaps to literary culture in general, through reduction in quality of published material. Training a Large Language Model on writers' output to generate ``sham books'' in a particular style seems to constitute a new form of plagiarism. This problem has
been little researched. In this study, we trained Machine Learning classifier models to distinguish short samples of human-written from machine-generated creative fiction, focusing on classic detective novels. Our results show that a Naïve Bayes and a Multi-Layer Perceptron classifier achieved a high degree of success (accuracy $> 95$\%), significantly outperforming human judges (accuracy $<55$\%). This approach worked well with short text samples (around 100 words), which previous research has
shown to be difficult to classify.
We have deployed an online proof-of-concept classifier tool, \textit{AI Detective}, as a first step towards developing lightweight and reliable applications for use by editors and publishers, with the aim of protecting the economic and cultural contribution of human authors.
}

\onecolumn \maketitle \normalsize \setcounter{footnote}{0} \vfill

\section{\uppercase{Introduction}}
\label{sec:introduction}

Generative AI has made remarkable advances in recent years, and is now widely available 
with the release of ChatGPT and similar systems. Along with many beneficial uses, there are diverse concerns for misuse, including generation of incorrect or harmful advice \cite{oviedo-trespalacios_risks_2023}, propagation of biases \cite{feng_pretraining_2023}, creation of deepfake video, fake news and fake product reviews \cite{botha_fake_2020}, and various forms of plagiarism, especially in scientific and other academic fields \cite{odri2023detecting}. Previous research has investigated techniques for identifying artificially generated text in these domains, with the aim of mitigating the societal harm from such misuse (see Section \ref{sec:bgroundprevious}).

Generative AI also presents a threat to the livelihood of writers and other creative artists, and may devalue their work. Models are often trained on writers' outputs, without their permission, and then the models can be used to generate similar content. 

\newpage
We  might \mbox{characterise} this problem as ``AI-mediated plagiarism'': rather than taking or modifying authors' work directly, a bad actor can create content using a generative AI trained on the authors' work. 
Increasing awareness of the issue is signalled by
developments such as the New York Times' announcement in 2023 that it was suing OpenAI and Microsoft for copyright infringement \cite{grynbaum_times_2023}.

We note a gap in the literature around developing detection tools in the area of creative fiction, and broach this problem by investigating whether Machine Learning (ML) models can reliably distinguish between short text samples from human-written novels and similar text automatically generated. 

Our early results show a good level of success, working with text excerpts from classic detective novels, and suggest that relatively simple classifiers can outperform humans in identifying automatically generated creative prose. Moreover, the approach works well with short text samples, which
previous studies found difficult to classify (see Section \ref{sec:comp-other}).

\section{\uppercase{Background}}
\label{sec:bground}

\subsection{Advances in Text Generation}
\label{sec:bgroundadvancements}

Recent increases in data availability and computing power have facilitated 
approaches to automatic text generation based on neural network models and deep learning \cite{goyal_systematic_2023}. 

A neural network is trained by optimising the weights and biases based on the observed \textit{vs.}\@ desired outputs \cite{bas_brief_2022}. As networks are not restricted to pre-existing patterns, the text generated by these models can be more ``creative'' according to the underlying semantic relationships \cite{pandey_natural_2023}. 

The introduction of Transformer models in 2017 represented another major advance \cite{vaswani2017attention}. The architecture of a Transformer consists of an encoder and a decoder. 
The encoder block contains a multi-head self-attention layer; the decoder block also has a cross-attention layer, enabling it to use the output of the encoder as context for text generation \cite{han_pre-trained_2021}. The attention component of Transformers underlies their success in text generation, as it enables a language model to decipher the correct meaning of a word using its context. For example, when ``it'' is used in a sentence, the model can better interpret what is being referred to. These types of text generation models learn from huge amounts of data, which enables them to generate high quality output. 

Increases in scale led to Large Language Models (LLMs), a notable early example being BERT from Google, which used the Transformer architecture to achieve dramatic improvements over earlier models \cite{devlin_bert_2019}. 

OpenAI's GPT-1 was developed in 2018, and version 3.5 was released to the public in 2022 via the ChatGPT interface, making this technology universally available. GPT-3 showed its ability to create text that was seemingly indistinguishable from human-written text \cite{pandey_natural_2023}. The initial version of GPT used 110 million learning parameters, and this number greatly increased with each version. GPT-2 used 1.5 billion and GPT-3 used 175 billion \cite{floridi_gpt-3_2020}. The number of parameters used for GPT-4 has been speculated to be approximately 100 trillion. Now ChatGPT has the ability to write human-like essays, news articles, and academic papers, as well as to complete text summarisation in multiple languages \cite{zaitsu_distinguishing_2023}. 

Many other companies have also launched LLMs with similar capabilities, including GitHub's and Microsoft's Copilot, Meta's LLama, Anthropic's Claude, and Google's Gemini.

\subsection{Text Generation -- Concerns for Misuse}
\label{sec:bgroundmisuse}

ChatGPT is freely available and widely adopted, raising the possibility of harm from inaccurate responses, or deliberate misuse by bad actors generating misleading texts.

One concern regarding text generated by LLMs is quality. As responses are based on statistical patterns and correlations found in large datasets, they can at times be irrelevant, nonsensical or offensive \cite{wach_dark_2023}; different models can vary regarding what is considered inaccurate or offensive \cite{feng_pretraining_2023}. As LLMs are pre-trained on large datasets which include opinions and perspectives, there is a risk of the introduction of biases for downstream models \cite{barclay_investigating_2024}. 
More ominously, ChatGPT can write fake news at scale, a task which was previously labour-intensive. This makes it easy to create media supporting or discrediting certain views, political regimes, products or companies \cite{koplin_dual-use_2023}. 

As awareness of these concerns increased, research has focused on ways to distinguish human-written from artificially generated text.

There has also been concern in creative industries that generative AI could be used to replicate artists' and writers' work, and possibly replace them. In 2023, members of the Writers Guild of America went on strike for 148 days seeking an agreement on protections regarding the use of AI in the television, film, and online media industries \cite{salamon_negotiating_2024}.

\subsection{Related Work}
\label{sec:bgroundprevious}

Numerous methods have been employed for the creation of AI-content detectors in other domains, including zero-shot classifiers, fine-tuning Neural Language Models (NLMs), 
as well as specialised classifiers trained from scratch \cite{jawahar_automatic_2020}.

OpenAI, the creator of ChatGPT, launched its own AI classifier to identify 
AI-generated text; however, they removed availability in July 2023 owing to its low accuracy (see https://platform.openai.com). A study which reviewed five AI-content detection tools (OpenAI, Writer, Copyleaks, GPTZero, and CrossPlag) observed high variability across the tools, and a reduced ability to detect content from a more recent version of GPT. Comparing GPT-4 with GPT-3.5 content, three of the five tools could find 100\% of GPT-3.5 content, but none achieved this level of detection for GPT-4 -- the highest result was 80\% by the OpenAI Classifier; then one detector managed 40\% and the rest only 20\%. This suggests that AI detection tools will have to evolve 
in response to the increasing sophistication of LLMs \cite{elkhatat_evaluating_2023}.  

The majority of zero-shot detectors evaluate the average per-token log probability of the generated text. A commercially available tool which uses zero-shot classification, GPTZero, has been found, in a study using medical texts, to have an accuracy of 80\% in identifying AI-generated texts \cite{habibzadeh_gptzero_2023}. 

Another example of a zero-shot detector is \mbox{``DetectGPT''}, which uses minor rewrites of text and then plots log probability of the original \textit{vs.}\@ the rewrites. The ``Perturbation Discrepancy'' created results in the log probability for human text tending towards zero, whereas the AI text was expected to give relatively larger values \cite{mitchell_detectgpt_2023}. DetectGPT was found to perform better than other existing zero-shot methods at detecting AI-generated fake news articles \cite{mitchell_detectgpt_2023}. However, a more recent study reports having outperformed DetectGPT by leveraging the log rank information \cite{su_detectllm_2023}.

Detectors using RoBERTa, which is based on the pre-trained model BERT, have been referred to as being ``state-of-the-art'' for AI text detection \cite{crothers_machine-generated_2023}.
The success of this approach is reported in a study which used a supervised Multi-Layer Perceptron (MLP) algorithm to train RoBERTa, achieving an accuracy of over 97\% on the test dataset. The dataset was created using URLs shared on Reddit, which were passed through GPT-3.5 Turbo for rephrasing. This study also achieved similar results using the Text-to-Text Transfer Transformer model (T5) as a starting point \cite{chen_gpt-sentinel_2023}. Another study based on an Amazon reviews dataset also found the RoBERTa model gave a 97\% accuracy, the highest result among those tested \cite{puttarattanamanee_comparative_2023}. 

An alternative approach for detecting whether text has been created by a human or an AI is by training a purpose-built Machine Learning model. One study looked at medical abstracts and radiology reports using several different ML techniques, including text perplexity, and singular and multiple decision trees. 
Perplexity is defined as the exponentiation of the entropy of the text,
giving an intuitive measure of the uncertainty in word choice.
This study also used one LLM, a pre-trained BERT model. 
Their results showed that ChatGPT created text which had a relatively lower perplexity (was more predictable) compared with human-written text. Nonetheless, perplexity gave the worst results compared with the other methods in this study, (with similar poor results for the singular decision tree). The multiple decision tree achieved a percentage F1 score of nearly 90\% for both datasets. The pre-trained BERT model still outperformed all other approaches with F1 scores 
above 95\% for both datasets \cite{liao_differentiate_2023}. 

Another study by Islam \textit{et al} tested eleven different Machine Learning models, with
a multiple tree model performing best with an accuracy of 77\%. The dataset for this study used a combination of news articles from CNN and data scraped from the question-and-answer website Quora \cite{islam_distinguishing_2023}. 

Support Vector Machine (SVM) models have demonstrated success in distinguishing between human-written and 
AI-generated text. A study which looked at identifying fake news on social media achieved 98\% accuracy, the highest result for all models tested within the experiments, using SVM algorithms \cite{sudhakar_detection_2024}. Another study, investigating SVM models to detect fake news, also achieved an accuracy of approximately 98\% when the title and first 1000 characters of the article were used in testing \cite{altman_detecting_2021}.

An experiment comparing logistic regression with a Naïve Bayes algorithm, where the dataset focused on identifying fake news, found the logistic regression algorithm performed better with an accuracy of 98.7\% \cite{sudhakar_effective_2022}. The use of statistical-based techniques has also been shown as a promising approach, such as ``GPT-Who'', which uses a logistic regression classifier to map extracted Uniform Information Density (UID) features. The hypothesis of UID is that humans prefer evenly 
spreading information without sudden peaks and troughs. This detector performed better than other statistical based detectors, and at the same level as fine-tuned transformer-based methods \cite{venkatraman_gpt-who_2023}. 

\subsection{Detection of Automatically Generated Creative Fiction}
\label{sec:related}

Prior studies addressed various domains including: medical-related text \cite{hamed_improving_2023,habibzadeh_gptzero_2023}, student submissions \cite{orenstrakh_detecting_2023,walters_effectiveness_2023,elkhatat_evaluating_2023}, and news articles \cite{islam_distinguishing_2023,mitchell_detectgpt_2023}, as well as popular websites with user contributions \cite{islam_distinguishing_2023,chen_gpt-sentinel_2023}. 

Despite concerns about the misuse of AI in the creative community, we note a considerable lack of research on detecting artificially generated creative fiction\footnote{We will use the term ``creative fiction'' as we do not distinguish ``genre'' fiction from ``literary'' fiction, but here we do not consider other forms of creative writing such as lyrics, screenplays, and poetry.}. A number of studies have investigated use of generative AI to \textit{assist} writers \cite{ippolito_creative_2022,landa-blanco_human_2023,guo_exploring_2024,gero_ai_2023,stojanovic_influence_nodate}. While this may raise literary questions regarding quality of writing, and philosophical considerations of the nature of human creativity, it does not directly threaten the livelihood of creative writers. Wholesale generation of entire novels is another matter. In 2024, the Authors' Guild noted the prevalence of ``sham books'' for sale on Amazon \cite {the_authors_guild_ai_2024}, and some high-profile stars signed an open letter requesting technology companies to pledge protection for human artists' work \cite{robins-early_billie_2024}. 

We do not expect that detectors trained on other domains will work well with creative fiction; indeed, one study demonstrated a 20\% reduction in effectiveness across different sources 
within a single domain (fake news) \cite{janicka_cross-domain_2019}.

The only study found in the literature addressing the detection of artificially generated creative fiction was the Ghostbuster study 
(which also included two other domains) 
\cite{verma_ghostbuster_2023}. However, this study did not use published novels.
The creative writing dataset was created from writing prompts and their associated stories from Reddit. When there was no prompt available for a story, first ChatGPT was given the story and asked to create a prompt, then that prompt was used to re-create the story. 

Ghostbuster is a sophisticated and effective model, created in three steps: each document was first fed into weaker language models to retrieve word probabilities, which were then combined into a set of features by searching over vector and scalar functions. The resulting features were used in a linear classifier. In the full Ghostbuster trial, the creative writing dataset performed well, but worse (F1 = 98.4\%) than two other more commonly used types of datasets, fake news and student 
essays (both F1 = 99.5\%). This suggests that it may be more difficult to distinguish AI-generated creative writing compared with other content types, underlining the need for further research in this area.

Additionally, this study attempted to address brittleness seen in other detectors' inability to generalise across different LLMs. Ghostbuster outperformed other models tested, but still showed a 6.8\% F1 decrease when analysing text from another LLM compared to ChatGPT. 

\subsection{Aims of this Research}
\label{sec:bgaims}

Having noted the concerns of the creative community, and the lack of research on detecting artificially generated creative fiction, we investigate the use of ML classifiers for this task, using only  short samples of text. We take this approach as:

\begin{itemize}
\item Earlier research has shown that ML models can be effective in detecting other types of AI-generated content.
\item As LLMs are rapidly evolving, and often proprietary, we do not wish to depend on the technology we are trying to detect.
\item A useful detector should be able to run independently with relatively low resources, so it could be deployed easily in the workflow of reviewers, editors, and publishers.
\end{itemize}

Additionally, we make a first comparison of the effectiveness of a ML-based classifier versus human judgment in identifying artificially generated prose.

\section{\uppercase{Experimental Method}}
\label{sec:method}

Our broad approach is to chop
human-written detective novels into a sequence of excerpts, then to generate similar texts using ChatGPT-3.5 Turbo, both by
rewriting existing excerpts and by using a customised prompt 
only (with no excerpt provided as an example).
The generated texts undergo just enough data preparation to ensure there are no obvious ``tells'' flagging their provenance. We then attempt to train ML classifier systems to distinguish the human-written from the machine-generated texts. We also compare the accuracy of the best classifiers with samples taken from two unseen novels, one by a different author, for insight into how well the models can generalise. We further compare the classifiers' success with that of 19 human judges who had attempted to make the same determination for a small selection of text samples in an online quiz.

We focus on short text samples, as these have proved difficult to detect in prior studies, and with a view to eventually creating a lightweight tool that could spot-check individual sections of text during the editing/publishing process.

\subsection{Datasets and Data Preparation}
\label{sec:methoddataset}

For the human-written prose, we used out-of-copyright novels by Agatha Christie from Project Gutenberg (https://www.gutenberg.org), as these are well-known, and the language is not too old-fashioned. Three novels\footnote{``The Murder on the Links'', ``Poirot Investigates'' \& ``The Man in the Brown Suit''.} were used in the base (human-written) dataset, creating 1424 excerpts, adding three more\footnote{``The Mysterious Affair at Styles'', ``The Big Four'' \& ``The Secret Adversary''.} in the extended base dataset (2713 excerpts).
Furthermore, one unseen Agatha Christie novel and one novel by another writer, Dorothy L Sayers\footnote{``The Secret of the Chimneys'' \& ``Whose Body?''}, were used to investigate 
the ability of the classifier to generalise to unseen but broadly similar text.

A Python script was used to create our base data set, by chopping the novels into a sequence of excerpts of the desired length, always terminating on a full stop\footnote{Terminating on other symbols such as ? and ! did not give reliable results.}, and removing extraneous text such as page numbers. In these experiments, all 
excerpts were of length ``approximately 100 words'', as described below. The texts were vectorised without pre-processing using Scitkit-Learn's CountVectorizer.

The AI-generated texts were produced in two ways: (1) by asking ChatGPT to rewrite a sample from  
the base dataset of novel excerpts, and (2) by asking it to write a similar text based on a prompt, with no sample text provided. Using a Python script, all requests were sent to OpenAI's API in a randomised order; one text excerpt was generated or rewritten per API request. 
We expect more robust results from the rewrites datasets, where we were able to generate many more
texts than in the prompt-only datasets, but we are still able to compare the two approaches.
For detection experiments, the generated texts in each dataset were mixed with the same number of human-written excerpts
randomly selected from a base dataset.

To enable a fair comparison of the models, 80\% of the base dataset was used as the training / test set -- this was split as 70\% training and 30\% test -- and the other 20\% was held back as an unseen validation set. 
The generated datasets used are summarised in Table \ref{tab:testsets}.

\begin{table*}[htbp]
\centering
\caption{Description of generated data sets.}
\label{tab:testsets}
\begin{adjustbox}{width=\textwidth}
\begin{tabular}{| p{2cm} | p{9cm} | p{2cm} | p{2cm} |}

\hline
\textbf{Data Set\newline Reference} & \textbf{Description} & \textbf{Type of AI generation} & \textbf{Total No. \newline examples} \\
\specialrule{.2em}{.1em}{.1em}
AC3Train & Training data separated out from a base data set created using 3 Agatha Christie books where human text was excerpts split out from the novels at approx. 100 word excerpts and the AI text was created using rewrites of the human text via the OpenAI API. & Rewrites & 1595 \\
\hline
AC3Test & Test data separated out from a base data set created using 3 Agatha Christie books where human text was excerpts split out from the novels at approx. 100 word excerpts and the AI text was created using rewrites of the human text via the OpenAI API.& Rewrites & 683 \\
\hline
AC3Unseen & Unseen data separated out from a base data set created using 3 Agatha Christie books where human text was excerpts split out from the novels at approx. 100 word excerpts and the AI text was created using rewrites of the human text via the OpenAI API.& Rewrites & 570 \\
\specialrule{.2em}{.1em}{.1em}
AC6Train& Training data separated out from a base data set created using 6 Agatha Christie books where human text was excerpts split out from the novels at approx. 100 word excerpts and the AI text was created using rewrites of the human text via the OpenAI API. & Rewrites & 3038 \\
\hline
AC6Test& Test data separated out from a base data set created using 6 Agatha Christie books where human text was excerpts split out from the novels at approx. 100 word excerpts and the AI text was created using rewrites of the human text via the OpenAI API. & Rewrites & 1302 \\
\hline
AC6Unseen& Unseen data separated out from a base data set created using 6 Agatha Christie books where human text was excerpts split out from the novels at approx. 100 word excerpts and the AI text was created using rewrites of the human text via the OpenAI API. & Rewrites & 1086 \\
\specialrule{.2em}{.1em}{.1em}
ChatGPTAC1 & ChatGPT was used with the prompt ``please write a story about a detective in the style of agatha christie'', after each response from ChatGPT, another prompt would be sent to ask for another story until there was enough text to create 10 text excerpts of approx. 100 words. For this dataset 10 text excerpts from the AC3 Data set were used as human text samples. & Prompt-Only (no text\newline provided) & 20 \\
\hline
ChatGPTGC1 & ChatGPT was used here to request a generic crime novel from the same time period (1920s) to be written. This time it was written in chapters and prompts were used to encourage ChatGPT to keep writing until enough text was available. This dataset has 12 approx. 100 word text excerpts from AI generation and again human text from another dataset (DLS1) was used to provide 12 human written samples. & Prompt-Only (no text\newline provided) & 24 \\
\specialrule{.2em}{.1em}{.1em}
DAC1 & Created using a different Agatha Christie Novel ``The Secret of Chimneys'' -- This dataset was created in the same way as the other datasets for this project. \newline The text was split into approx. 100 word chunks to the closest full sentence and the text excerpts were sent through the OpenAI API to be re-written. 100 random samples were extracted from the original human set and from the results from the API. & Rewrites & 200 \\
\hline
DLS1 & Whose Body? A Lord Peter Wimsey Novel by Dorothy L. Sayers (1923) -- This is a novel from a crime author which was written around the same time as the Agatha Christie novels. Again 100 samples from the original human text were used along with 100 results from the OpenAI API, completely randomised. & Rewrites & 200 \\
\hline

\end{tabular}
\end{adjustbox}
\end{table*}

\subsection{OpenAI API Settings}
\label{sec:methodsopenai}

The temperature setting and prompt wording used for the AI-generated text were derived by extensive trial-and-error over many iterations, arriving at a prompt which produced text with no obvious tells that it had been artificially generated. We noted that
too long or specific a prompt sometimes resulted in some requirements 
being ignored.
Lower temperatures generated less varied text; in particular, rewritten excerpts often just had a few word substitutions. Too high a temperature resulted in text too unlike the target material. 

The final OpenAI API settings are shown below: 

\begin{itemize}
\item model=``gpt-3.5-turbo-0125''
\item temperature=0.7
\item prompt = \textit{``You will take the role of an author of crime novels. A text excerpt will be provided, you have to review it for number of space characters and key details. Create a new text excerpt which contains the same key details but appears structurally different to the original. The new text must have approximately the same number of spaces as the original. Only return the new text passage. Do not include place holders, line breaks or any other text except the new passage. Text excerpt:''}
\end{itemize}

\subsection{Text Excerpt Length Distribution}
\label{sec:resultsdatasetlength}

\begin{figure}[htbp]
\centering 
\includegraphics[scale=0.5]{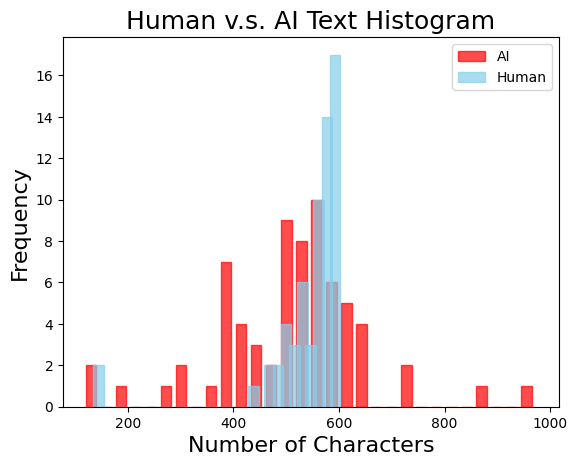}
\caption{Distribution of character lengths for Human-written and AI-generated text.}
\label{fig:huma_vs_ai_text_org}
\end{figure}

The lengths of excerpts in the base datasets vary somewhat due to the 
requirement of separating chunks at a full stop.
The OpenAI API did not prove accurate in generating texts of a required length, although there was some modest improvement by requesting a particular number of spaces, rather than number of words. Moreover, the API showed a preference towards generating shorter texts rather than longer. Figure \ref{fig:huma_vs_ai_text_org} shows how the distribution of character length is significantly different between the base and generated datasets when the expectation was to target approximately 600 characters (100 words).

With a difference in mean length of 42 characters, and a large
jump in standard deviation (68 \textit{vs.}\@ 120),
the length distribution raised a problem as it could bias the classification.
This issue was addressed in two steps.
For the human-written text preparation, the text was cut (at the nearest full stop) to a randomly 
selected length 
within a defined range; and for all datasets, outliers were removed to reduce the variations in length.
The resultant ``balanced'' datasets had excerpts of the same mean length
(563 characters) and much closer standard deviations (61 \textit{vs.}\@ 81).
All experiments reported here were run on these balanced datasets.

\subsection{Models Tested}
\label{sec:methodsmodels}

We tested six ML models from Scikit Learn (https://scikit-learn.org), based on their earlier use in detecting generated text in other domains. These were: Support Vector Machine, Logistic Regression, Random Forest, MLP Classifier, Decision Tree and Naïve Bayes. The inclusion of a single decision tree is to allow comparison with the Random Forest (multiple decision trees), and the decision to include Naïve Bayes was based on its known ability in text classification tasks, where it is computationally efficient and often exhibits a good predictive performance \cite{chen_feature_2009}. 

\begin{table*}[htbp]
\caption{Model comparison using six models and three-novel datasets. 
}
\label{tab:modelcomp}
\centering
\begin{adjustbox}{width=\textwidth}
\begin{tabular}{| l | l | l | l | l |}
\hline
\textbf{Model} & \textbf{AC3Test Accuracy} & \textbf{AC3Unseen Accuracy} & \textbf{AC3Test F1 Score} & \textbf{AC3Unseen F1 Score} \\
\specialrule{.2em}{.1em}{.1em}
Naïve Bayes Multinomial & 93.42\% & 92.98\% & 92.63\% & 92.39\% \\
\hline
SVM & 92.84\% & 92.52\% & 91.75\% & 91.77\% \\
\hline
Logistic Regression & 92.98\% & 92.73\% & 91.40\% & 91.54\% \\
\hline
Decision Tree & 71.93\% & 71.00\% & 76.49\% & 77.59\% \\
\hline
MLP Classifier & 94.74\% & 94.50\% & 92.81\% & 92.67\% \\
\hline
Random Forest & 88.01\% & 86.90\% & 90.00\% & 89.84\% \\
\hline
\end{tabular}
\end{adjustbox}
\end{table*}

\begin{table*}[htbp]
\caption{Results from the two top-performing models  following optimisation and trained using data from three 
novels.}
\centering
\label{tab:extestsets}
\begin{tabular}{| l | l | l | l | l | l |}
\hline
\textbf{Dataset} & \textbf{Model} & \textbf{Accuracy} & \textbf{Precision} & \textbf{Recall} & \textbf{F1} \\
\specialrule{.2em}{.1em}{.1em}
AC3Test & MLP Classifier & 95.03\% & 96.89\% & 92.86\% & 94.83\% \\
\hline
AC3Test & Naïve Bayes & 93.86\% & 97.12\% & 90.18\% & 93.52\% \\
\hline
AC3Unseen & MLP Classifier & 92.28\% & 94.14\% & 90.18\% & 92.11\% \\
\hline
AC3Unseen & Naïve Bayes & 92.28\% & 94.80\% & 89.47\% & 92.06\% \\
\specialrule{.2em}{.1em}{.1em}
ChatGPTGC1 & MLP Classifier & 95.83\% & 92.31\% & 100.00\% & 96.00\% \\
\hline
ChatGPTGC1 & Naïve Bayes & 100.00\% & 100.00\% & 100.00\% & 100.00\% \\
\hline
ChatGPTAC1 & MLP Classifier & 85.00\% & 81.82\% & 90.00\% & 85.71\% \\
\hline
ChatGPTAC1 & Naïve Bayes & 95.00\% & 90.91\% & 100.00\% & 95.24\% \\
\hline
\end{tabular}
\end{table*}

\subsection{Experimental Runs}
\label{sec:xruns}

An initial experiment on all six models with the three-novel datasets identified the MLP and the Naïve Bayes models as the best performing. These two models were then optimised and carried forward for testing against all the AI-generated datasets, with a follow-on run for these two models where the training data was increased to six novels. Lastly, to investigate generalisability of the classifiers, we tested them against excerpts from two previously unseen novels, 
one by Agatha Christie, and one of similar style from a different author, Dorothy L Sayers.

\section{\uppercase{Experimental Results}}
\label{sec:expres}

\subsection{Model Comparison}
\label{sec:rdiscussionmodelcomp}

Table \ref{tab:modelcomp} summarises results comparing the models trained on the AC3Train dataset and tested using the AC3Test and AC3Unseen datasets. 

The MLP Classifier and Naïve Bayes models performed best overall, while
the SVM and Logistic Regression models also gave good results.
As expected, the decision tree model performed the worst with all results lower than 80\%; the random forest model performed better, but still significantly below the other models.

Based on the results for the AC3Test dataset, it would be expected that the MLP Classifier would perform better than the Naïve Bayes for the AC3Unseen dataset. The results, however, were the same for both models. On inspection, it was found that generally the same samples in the AC3Unseen set were mislabelled by both models.  

The MLP Classifier and Naïve Bayes models were tuned and carried forward for further experimentation.
Of the many
tunable hyper-parameters of the MLP model, the only change that improved accuracy was setting the hidden\_layer\_sizes to one layer with 155 units: this enabled the accuracy to exceed the 95\% mark and the F1 score was slightly improved also.
For the Naïve Bayes model, the highest accuracy for the AC3Test data set was achieved by the multinomial algorithm using an alpha value of 0.7.

The results of the tuned models are shown in Table \ref{tab:extestsets}. Both classifiers show similar results on the rewrites
datasets AC3Test and AC3Unseen, with more variation in the results for the prompt-only
datasets ChatGPTGC1 and ChatGPTAC1. We note that for rewritten texts, precision exceeds recall, 
while the inverse obtains for the the prompt-only datasets. This implies some qualitative
difference resulting from the two methods used for generating the texts with GPT.

\subsection{Six-Novel Data Experiments}
\label{sec:6nov}

The results of the experiments using increased training data with the optimised models are shown in Table \ref{tab:extestresults}.
For the Naïve Bayes model, all but one test set saw an increase in scores with the six-novel dataset AC6train, and for the MLP Classifier model, an improvement in several tests was also observed.

The larger training dataset significantly improved the discrimination of both models. The average accuracy across all tests for the MLP Classifier is now 96.09\% (previously 92.76\%) compared with the Naïve Bayes which now has an average accuracy of 96.05\% (previously 94.34\%). Looking at the percentage F1 scores, the MLP Classifier achieved an average of 96.02\% (previously 92.74\%) compared with the Naïve Bayes which now has an average F1 score of 95.94\% (previously 94.11\%).
Both classifiers achieved 100\% accuracy for the ChatGPTAC1 dataset, rising from
85\% and 95\% previously.
Although this is still a small test dataset, reaching the 100\% score as a result of doubling the training data gives some confidence in the result.

However, two datasets did show minor reductions in accuracy. The only model whose
score was reduced by more then 1\% was Naïve Bayes, falling to 95.83\% on the
prompted ``generic crime novel'' dataset ChatGPTGC1. This may be a result of overfitting on 
the previous smaller dataset.
However, the overall improvements with
the larger training dataset show it is clearly beneficial overall.

\subsection{Generalisation Experiments}
\label{sec:genexp}

Testing against previously unseen novels gave encouraging results, shown in Table \ref{tab:diffnovelstestsets}.
For the DAC1 dataset (an unseen novel by the same author), accuracy was over 90\%
on all runs; the MLP Classifier performed better than Naïve Bayes, with recall being the main issue for both models.
For the DLS1 dataset (an unseen novel by a different author), the results were comparable or better for both models compared with seen novels. 
For the MLP Classifier, the accuracy of the DLS1 dataset at 95.41\% is extremely close to that of the AC6Unseen set (95.67\%). For the Naïve Bayes model, the DLS1 dataset at 95.92\% accuracy actually performs better than the AC6Unseen set (95.03\%). 
Although this higher score on an unseen novel may not be significant, it is clear that the models can generalise well, at least over the same genre of creative fiction. 

\begin{table*}
\centering
\caption{Results from the two top-performing models trained using data created from six 
novels (increased training data).}
\label{tab:extestresults}
\begin{tabular}{| l | l | l | l | l | l |}
\hline
\textbf{Dataset} & \textbf{Model} & \textbf{Accuracy} & \textbf{Precision} & \textbf{Recall} & \textbf{F1 } \\
\specialrule{.2em}{.1em}{.1em}
AC6Test & MLP Classifier & 94.62\% & 95.97\% & 92.98\% & 94.45\% \\
\hline
AC6Test & Naïve Bayes & 95.01\% & 98.48\% & 91.26\% & 94.74\% \\
\hline
AC6Unseen & MLP Classifier & 95.67\% & 97.51\% & 93.74\% & 95.59\% \\
\hline
AC6Unseen & Naïve Bayes & 95.03\% & 97.48\% & 92.45\% & 94.90\% \\
\specialrule{.2em}{.1em}{.1em}
ChatGPTGC1 & MLP Classifier & 95.83\% & 92.31\% & 100.00\% & 96.00\% \\
\hline
ChatGPTGC1 & Naïve Bayes & 95.83\% & 92.31\% & 100.00\% & 96.00\% \\
\hline
ChatGPTAC1 & MLP Classifier & 100.00\% & 100.00\% & 100.00\% & 100.00\% \\
\hline
ChatGPTAC1 & Naïve Bayes & 100.00\% & 100.00\% & 100.00\% & 100.00\% \\
\hline
\end{tabular}
\end{table*}

\begin{table*}[htbp]
\caption{Experimental results from the two top performing models on unseen novels.}
\centering
\label{tab:diffnovelstestsets}
\begin{tabular}{| l | l | l | l | l | l | l |}
\hline
\textbf{Dataset} & \textbf{Training\newline Set} & \textbf{Model} & \textbf{Accuracy} & \textbf{Precision} & \textbf{Recall} & \textbf{F1} \\
\specialrule{.2em}{.1em}{.1em}
DAC1 & AC3Train & MLP Classifier & 92.50\% & 98.85\% & 86.00\% & 91.98\% \\
\hline
DAC1 & AC3Train & Naïve Bayes & 90.50\% & 98.80\% & 82.00\% & 89.62\% \\
\hline
DLS1 & AC3Train & MLP Classifier & 95.92\% & 98.91\% & 92.86\% & 95.79\% \\
\hline
DLS1 & AC3Train & Naïve Bayes & 94.39\% & 96.77\% & 91.84\% & 94.24\% \\
\hline
DAC1 & AC6Train & Naïve Bayes & 94.50\% & 100.00\% & 89.00\% & 94.18\% \\
\hline
DAC1 & AC6Train & MLP Classifier & 95.00\% & 98.91\% & 91.00\% & 94.79\% \\
\hline
DLS1 & AC6Train & Naïve Bayes & 95.92\% & 97.87\% & 93.88\% & 95.83\% \\
\hline
DLS1 & AC6Train & MLP Classifier & 95.41\% & 97.85\% & 92.86\% & 95.29\% \\
\hline
\end{tabular}
\end{table*}

\subsection{Model Runtimes}
\label{sec:resultsmodelrun}

The time for completion for the Naïve Bayes model was 5 seconds for the original training/test/validation set and 8 seconds for the increased training dataset. However, the MLP Classifier took considerably longer with 38 seconds for the original set and 61 seconds for the larger set. 

As the results of the two optimised classifiers were close in terms of accuracy and F1 scores, the much faster run time for the Naïve Bayes model recommends this as the better selection for the classification tool described in Section \ref{sec:classifiertool}. 

\subsection{ChatGPT-4o}
\label{sec:ver4}

Since our original experiments, ChatGPT-4 has been released. We conducted a mini-experiment, using an unchanged training set and new test sets of randomly selected novel excerpts. There was a slight drop in performance when using ``gpt-4o-mini'' (average accuracy 94.25\% compared with the original 95.03\%). A larger drop was observed using ``GPT-4o'' (average accuracy 89.25\%). We will conduct more extensive experiments, but we note the drop in performance is modest given the huge number of parameters in GPT version 4.

\section{\uppercase{Detection by Human Judges}\label{sec:human}}

To determine whether the excerpts generated by ChatGPT could easily be detected upon reading, a small quiz was set up in Google Documents. This displayed 10 text excerpts that were either taken from the human-written novels or the AI-rewritten text. A total of 19 people completed the quiz, which is available here: https://forms.gle/JhApKWkC9CAHXRmo8. Only 10 examples were included, to encourage completion, as the main purpose of the quiz was just to ensure no obvious tells had been overlooked during data preparation, allowing too-easy classification of the texts. A larger survey would be required for robust analysis; nonetheless, we see some interesting results.

\begin{figure}[htbp]
\centering  
\includegraphics[scale=0.35]{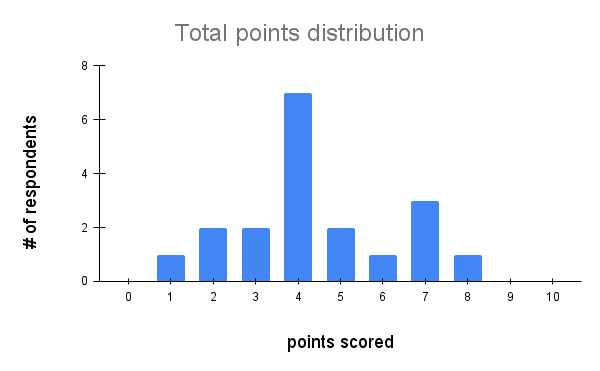}
\caption{Distribution of points scored on a Human \textit{vs.}\@ AI quiz}
\label{fig:user_eval}
\end{figure}

The judges' scores were normally distributed with a median score of 4 and a mean of 4.42, 
as shown in Figure \ref{fig:user_eval}.
Thus, the humans showed little ability to identify the AI-generated texts. 

With only 19 judges, we do not have sufficient data to determine if the results are no better than chance, but a one-tailed \mbox{t-test} provides good evidence that the respondents' ability to distinguish
between the text samples
is under 55\% in accuracy $(t(18) = -2.5$, p-value $= 0.01$, 95\% percent confidence interval: $-\infty:52\%$), and it is certainly far below the ability of the ML classifiers. 

These results are in line with the literature: for example, Clark \textit{et al} report that
untrained human evaluators score no better than chance trying to identify text from ChatGPT-3 in other domains,
with results rising to 55\% with practice \cite{clark_all_2021}.

\section{\uppercase{Classifier Tool}}
\label{sec:classifiertool}

The Naïve Bayes model was used to create an online classifier tool,
\textit{AI Detective},
for general experimentation. The tool is available at: https://tinyurl.com/ai-detective.
This tool used the AC6Train data for training, and can accept unseen test data and attempt to classify the input as human-written \textit{vs.}\@ AI-generated. The tool can also be used to experiment with different training and/or test sets: instructions for use are provided within the Google Colab script. All our code and data are available for replication and experimentation from the links at this URL.

\section{\uppercase{Discussion}}
\label{sec:disc}

\subsection{Summary and Review of Findings}

Of the six Machine Learning models tested, four gave promising results with the top two models optimised and evaluated with multiple test sets. The results from the best performing models, Naïve Bayes and MLP Classifier, show that it is possible to detect AI-generated creative writing with high accuracy using only short text samples. The accuracy scores for the Naïve Bayes model ranged from 94.5\% to 100\%, with an average of 96.05\%. The MLP Classifier performed with similar accuracy, ranging from 94.62\% to 100\% with an average of 96.09\%. However, the Naïve Bayes model was much more efficient, running in 8 seconds compared with over 1 minute. The average percentage F1 score across all test sets for the Naïve Bayes model was 95.94\% with a range of F1 scores from 94.18\% to 100\%. 

The larger reduction in scores for the MLP Classifier on the ChatGPTAC1 dataset was unexpected as this model had been performing well. This suggests that the training data may be more tightly fitted on the MLP Classifier model, and therefore it cannot generalise as well as the Naïve Bayes model. 

Looking at the recall results, the Naïve Bayes model was able to identify all instances of AI text generated directly from prompts (\textit{i.e.}\@ not rewritten), and the MLP Classifier only missed one sample overall.
For both Naïve Bayes and MLP Classifier algorithms, it was more common for AI-generated text to be incorrectly predicted as human text than \textit{vice versa}.
Comparing the metrics for the rewritten versus the prompt-only test datasets, precision and recall show opposite trends.
For text generated from scratch, the models have higher recall and are therefore identifying more instances of AI text; however, in the rewrites datasets, the precision is higher. We speculate that the rewrites may still mirror elements of human text, making them harder to identify. 

The results from both top models show that they can generalise well to other novels from the same time period and genre. For the MLP Classifier, the accuracy of the DLS1 dataset, based on the Dorothy L Sayers novel, was 95.41\%, very close to that of the AC6Unseen set at 95.67\%. For the Naïve Bayes model, the DLS1 dataset at 95.92\% outperforms the AC6Unseen set at 95.03\% accuracy. How well this generalisation can extend to other styles of writing remains to be investigated. 

\subsection{Comparison with Other Work}\label{sec:comp-other}

One study utilising a Naïve Bayes model achieved an accuracy of 94.85\% using fake political news data \cite{sudhakar_effective_2022}, somewhat better than our AC3 datasets, but outperformed by our AC6
datasets. Moreover, our results for the MLP classifier are superior to a previous study which only achieved 72\% \cite{liao_differentiate_2023}, where the dataset was created from news and social media content. It would be interesting to investigate whether different classifiers perform better with particular styles of writing.

Ghostbuster, the only other study to include datasets derived from creative fiction, 
achieved an F1 score of 98.4\% in this domain \cite{verma_ghostbuster_2023}. While this is higher than the F1 scores reported here, 
Ghostbuster's score is over texts of all lengths. Their in-domain F1 score
drops to around 85\% for texts of 100 tokens, and the authors state that performance ``may be unreliable for documents with $\le$ 100 tokens''.
We note also that OpenAI recommended that AI detectors should use a minimum of 1000 characters for reliability (see https://platform.openai.com).  
Our results suggest therefore that our ML classifier models may perform better on short text samples, at least within one style of creative fiction.

Moreover, the Ghostbuster study did not use published novels, and their approach was more similar to the prompt-only datasets tested here. Our average F1 score for the prompt-only datasets, shown in Table \ref{tab:extestresults}, reached 98\% and therefore is almost as high as Ghostbuster,
despite their using texts of greater average length.

\subsection{Limitations of this Study}
\label{sec:limits}

In this pilot study, the ML classifiers have been trained on only one author, and mostly tested on the same
author, with limited testing on one other author from the same genre and time period. 

Moreover, although many LLMs are now available, only ChatGPT has been used for the AI-generated texts.

Experiments with more data would increase the robustness of our results, especially for the prompt-only datasets. Furthermore, achieving an effective prompt was harder than expected, and further prompt engineering may be required to improve the quality
of the AI-generated texts. 

Owing to the difficulties in generating texts of a stipulated length, there
is still some difference in the variance of text length in the
datasets used, and other data-preparation artefacts may remain which could aid the classifiers. 
For example, as ChatGPT was requested to review the details of each text passage and use 
it to create a new one, the AI-generated rewrites may appear more ``self-contained'' 
compared with the chunked off parts of a novel appearing in the base datasets. 

We noted 
also that the AI-generated rewrites sometimes introduced named entities that were not 
present in the original text.  Such artefacts of the data preparation process may have 
assisted the classifiers in identifying the automatically generated texts.

At present we have limited evidence that the classifiers will generalise 
to other, similar styles of writing, but it is unclear how far a classifier can generalise
before retraining would be required.
These issues will be addressed in future work. 

\section{\uppercase{Future Work and Conclusion}}
\label{sec:fwconc}

Here we present only preliminary results. It remains to run further tests with more systematically generated ChatGPT prompts, with larger prompt-only datasets, with texts of different lengths, and with further tuning of the models. We intend especially to investigate the effect of gradually generalising our unseen test data sets to creative fiction increasingly different in style from the training data.

Both the MLP and Naïve Bayes classifiers achieved better accuracy with the prompt-only test data, compared with the rewritten AC6Unseen dataset. While this may be an artefact of the differing sizes of the datasets, the implication is that it may be easier for the classifiers to identify wholly generated rather than rewritten text, which we may expect to mirror the
human-authored excerpts more closely. This is worthy of further investigation. 

Expanded experiments with human judges are also warranted, collecting more data to establish significance, and with attention to what characterises excerpts that are particularly easy/hard to detect, and to whether particular individuals (perhaps avid readers) have a higher than average ability to distinguish human-written from AI-generated text. A more developed version of the classifier tool may support training humans in this skill.
 
Nonetheless, our early results indicate that it is possible for a low-resource classifier to detect artificially generated creative fiction with a high degree of accuracy, based only on a short sample of the text. This opens the door to the construction and deployment of easily used tools for editors and publishers, helping to protect the economic and cultural contribution of human writers in the age of generative AI.

\bibliographystyle{apalike}
{\small
\bibliography{references}}

\end{document}